\documentclass{article}
\usepackage{spconf,amsmath,graphicx,hyperref}
\usepackage{booktabs}
\usepackage{graphicx}
\usepackage{subcaption}
\usepackage{makecell}
\usepackage{enumitem}
\setlist[itemize]{leftmargin=*, topsep=0pt, itemsep=0pt, partopsep=0pt, parsep=0pt}

\newcommand{\remark}[1]{}

\newcommand{\full}[1]{}                 

\begin{document}
\title{Webscraper: Leverage Multimodal Large Language Models for Index-Content Web Scraping}

\name{Guan-Lun Huang
 \quad
Yuh-Jzer Joung 
}
\address{Dept.~of Information Management, National Taiwan University, Taipei, Taiwan}

\maketitle

\begin{abstract}
Modern web scraping struggles with dynamic, interactive websites that require more than static HTML parsing. Current methods are often brittle and require manual customization for each site. To address this, we introduce Webscraper, a framework designed to handle the challenges of modern, dynamic web applications. It leverages a Multimodal Large Language Model (MLLM) to autonomously navigate interactive interfaces, invoke specialized tools, and perform structured data extraction in environments where traditional scrapers are ineffective. Webscraper utilizes a structured five-stage prompting procedure and a set of custom-built tools to navigate and extract data from websites following the common ``index-and-content'' architecture. Our experiments, conducted on six news websites, demonstrate that the full Webscraper framework, equipped with both our guiding prompt and specialized tools, achieves a significant improvement in extraction accuracy over the baseline agent Anthropic's Computer Use. We also applied the framework to e-commerce platforms to validate its generalizability.


\end{abstract}


\section{Introduction}
The exponential growth of websites and online data has made web crawling a critical mechanism for supplying data. As large-scale applications—such as the pre-training of Large Language Models (LLMs)—continue to expand, the need for timely and accurate web scraping has become increasingly essential.
News websites, in particular, are valuable due to their immediacy, supporting applications such as public opinion analysis that inform decision-making for governments, enterprises, and organizations.

However, web crawling on modern news sites remains challenging. Unlike early static pages where content was embedded in HTML, contemporary websites often load information dynamically through JavaScript (e.g., infinite scrolling) and include noise such as advertisements and scripts. Current expert-designed crawlers are brittle, requiring significant manual effort and frequent updates when websites change their structure.

To address these issues, we propose a framework that leverages Multimodal Large Language Models (MLLMs). By processing webpage screenshots, MLLMs act as if they possess visual perception. Coupled with browser-control tools simulating mouse and keyboard interactions, they can navigate websites dynamically in a human-like manner. To complement this, MLLMs are also equipped with HTML parsing tools, ensuring that hidden or dynamically loaded content can be extracted reliably. The MLLM autonomously decides when to use browsing tools and when to employ parsing tools, based on the user’s query and the current webpage state.

Our contribution is three-fold:
\begin{itemize}
    
    \item We demonstrate the feasibility of MLLM-based prompt design for index-content style web scraping tasks.
    
    \item We enhance task accuracy by integrating MLLMs with specialized extraction tools.
    
    \item We evaluate the generalization of the proposed framework to other domains, such as e-commerce web scraping.
\end{itemize}

\begin{figure*}[t]
    \centering
    \includegraphics[width=0.8\textwidth]{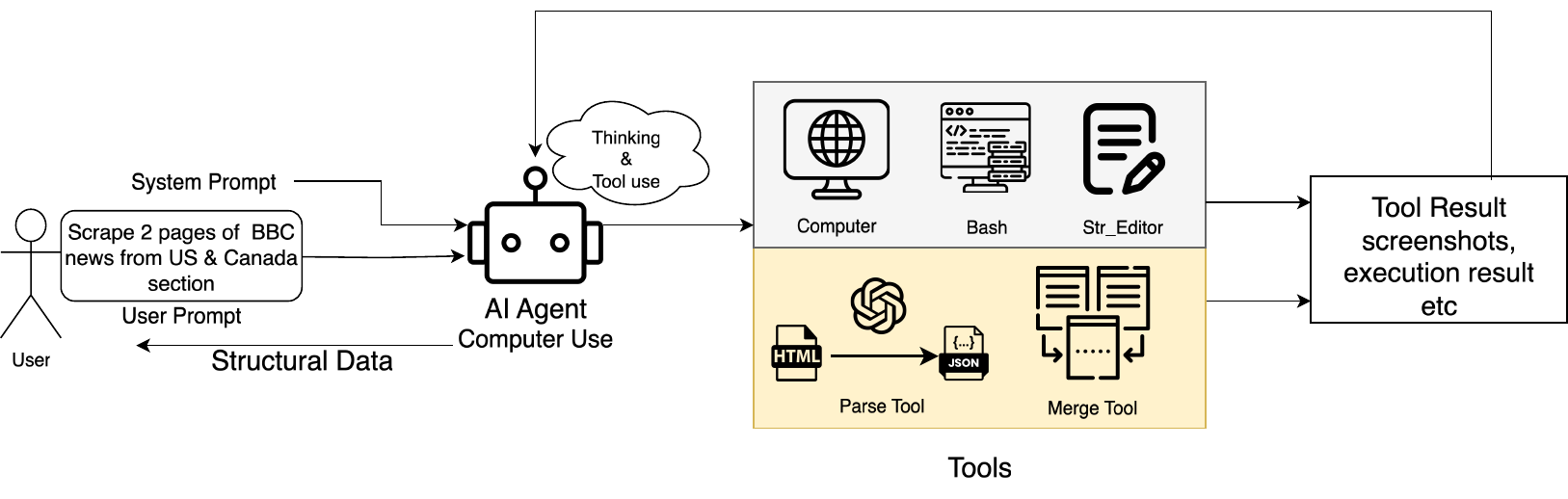}
    \caption{An overview of our proposed system architecture. The agent (Computer Use) receives prompts and uses a combination of native environment tools (top right) and our custom-built scraping tools (bottom right) to perform the extraction task and produce structured data.}
    \label{fig:system_arch}
\end{figure*}

\section{Related Work}

Traditional web scraping techniques, relying on static HTML parsing, are increasingly ineffective on modern websites built with dynamic JavaScript frameworks. These sites require user interaction to render content, necessitating a more sophisticated, two-stage approach: (1) Web Navigation, where an autonomous agent interacts with the interface to surface information, followed by (2) Web Information Extraction (WIE), where the revealed content is parsed from semi-structured HTML documents. 

In Web Navigation, previous work has leveraged MLLMs to create powerful agents. Systems like SeeAct~\cite{pmlr-v235-zheng24e} and WebVoyager~\cite{he-etal-2024-webvoyager} use visual understanding to automate tasks within the browser, but their capabilities diminish significantly in more complex desktop environments~\cite{10.1007/978-3-031-73113-6_10}. To address this, the frontier of agent design has shifted towards more general reasoning abilities. This is exemplified by frameworks like Anthropic's Computer Use~\cite{ClaudeComputerUse}, which employs an iterative Observe-Reason-Act loop~\cite{yao2023reactsynergizingreasoningacting} to give agents the fine-grained control needed to handle a wider array of complex, interactive tasks.

Concurrently, the field of Web Information Extraction (WIE) has evolved from brittle, rule-based wrappers~\cite{Kushmerick1997WrapperIF,article,10.1145/301136.301191}. While modern LLM-based approaches\cite{10.1145/3485447.3512032,lockard-etal-2020-zeroshotceres,wang-etal-2023-mustie} achieve high accuracy, their high computational cost presents a practical hurdle for large-scale scraping. To address this efficiency problem, a recent trend focuses on using an LLM once to generate a reusable, low-cost scraper~\cite{huang2025automatic,li2024xpathagentefficientxpath,huang-etal-2024-autoscraper}. AutoScraper~\cite{huang-etal-2024-autoscraper} is a prime example of this paradigm, employing an LLM to progressively generate and validate a set of robust XPath expressions. This XPath-generation approach, while cost-effective, is still fundamentally geared towards locating individual data points. It is not inherently designed for the holistic extraction of large-scale structured data, such as capturing all related entries in a product catalog from a single page.


In short, while prior work has produced powerful agents for fine-grained interaction and more efficient models for targeted data extraction, there remains a need for a unified framework that leverages advanced navigational capabilities to robustly manage the dynamic and interactive nature of modern websites. Addressing this gap is the central goal of our research.

\section{Methodology}


\subsection{Task Definition: Index-and-Content Scraping}


Our framework targets the widely used \emph{index-and-content} web architecture, which underlies many familiar websites, including news portals (listing articles), e-commerce platforms (displaying product search results), video platforms (showing video galleries), and social media feeds (aggregating posts). It consists of two primary components:

\begin{itemize}
\item \textbf{An Index Page}: A central page that serves as a directory, listing multiple items (e.g., a news category, a product search result). Each item typically contains a hyperlink.
\item \textbf{Content Pages}: The individual destination pages reached by clicking the links on the index page, containing the detailed information for each item (e.g., a full news article, a product description page).
\end{itemize}

The scraping task is initiated by a single, natural-language user prompt that defines the entire objective. This prompt specifies the target website, the scope of the crawl, and the data fields to be extracted. For example:

\textit{``Scrape the first two pages of US-Canada section from BBC news https://www.bbc.com/news/us-canada, extracting the title, link, content for each news.''}

The expected output of a successful task is a single, structured JSON file containing a list of objects, where each object represents an extracted item with its corresponding data fields.

\subsection{System Architecture}

Our goal is to develop a robust framework that transforms a general-purpose GUI agent into a specialized, autonomous web scraper. While general agents excel at interaction, they often lack the procedural knowledge needed for large-scale, structured data extraction. Therefore, we enhance a foundational agent (Computer Use) with two key components: (1) specialized tools for parsing and merging data, and (2) a guiding system prompt that coordinates a structured, five-stage extraction process. As shown in Figure~\ref{fig:system_arch}, our system integrates a foundational agent with two distinct sets of tools—native and custom—all orchestrated by a central system prompt.



\subsubsection*{Foundational Agent: Computer Use}
We use Anthropic's Computer Use framework as our base layer. It provides an MLLM-driven agent that receives a user prompt (e.g., ``Scrape 2 pages of BBC news'') and a system prompt, then intelligently selects tools to accomplish the goal.

\subsubsection*{Native Environment Tools}
The agent has access to a set of native tools that allow it to interact with a standard computer environment:

\begin{itemize}
    \item Computer: This tool enables GUI-level interactions, such as viewing the screen, moving the mouse, clicking, and scrolling. It serves as the agent's primary means of navigating web pages like a human user.
    \item Bash: Provides a command-line interface for file system operations (e.g., mkdir, ls), running scripts, and network requests (e.g., curl).
    \item Str\_Editor: A tool for programmatically writing, reading, and modifying files. The agent uses this to generate and refine extraction scripts.

\end{itemize}


We extend the agent's capabilities by introducing two custom tools, shown in the yellow-highlighted area of Figure 1:
\begin{itemize}
    \item Parse Tool: This tool converts raw HTML into structured data by delegating, or ``outsourcing,'' the entire parsing task. When invoked, it sends the HTML and user requirements to a more powerful reasoning model (GPT-o3), which generates a tailored Python script. This script is then executed within a GPT code interpreter environment to produce the structured data. This delegation strategy is a deliberate design choice with three key advantages:
    \begin{itemize}
        \item It preserves the main agent's context window by offloading the verbose code generation and execution logic.
        \item 
It enhances robustness by delegating the mission-critical task of index-page parsing to a more powerful, specialized model, thereby ensuring accurate initial data collection.
        \item It simplifies the system prompt, allowing it to focus on high-level strategic guidance rather than low-level parsing instructions.
    \end{itemize}
    \item Merge Tool: This tool aggregates lists of structured data collected across multiple scraping iterations. It handles deduplication and consolidates results, which is crucial during iterative processes like handling pagination.
\end{itemize}

\section{Experiments}

\begin{remark}{
This section details the empirical evaluation of our proposed framework. Our experiments are designed to answer two primary research questions:

\begin{itemize}
    \item  RQ1 Does a structured, prompt-guided approach provide significant performance gains over a naive, baseline application of a general-purpose agent for web scraping?
    \item  RQ2 Does the integration of functional, external tools provide a measurable performance benefit over a purely prompt-based strategy?
    \item  RQ3 To what extent does the Webscraper framework generalize to diverse domains, such as e-commerce, that also exhibit the index-and-content architecture?
\end{itemize}

To answer these questions, we designed a comparative study of three distinct methods.
}\end{remark}


\subsection{Experiment Setting}

To evaluate our framework, we consider the following three configurations:\footnote{All source code and prompts are placed in~https://github.com/chinaoel/webscraper.}
\begin{itemize}
    \item Baseline Agent: This represents a zero-shot application. The user's scraping task is passed directly to the default Computer Use agent (claude-3-7-sonnet-20250219) without our specialized system prompt or custom tools.
    \item Webscraper (Prompt Only): An ablation setting to isolate the effect of prompting.  The baseline agent is augmented with our guiding system prompt, but the custom Parse Tool and Merge Tool are disabled. Instead, descriptions of the tool operations are embedded in the prompt to guide the agent.
    \item Webscraper (Prompt + Tool): Our full proposed framework, which equips the agent with both the guiding system prompt and the functional Parse Tool and Merge Tool.
\end{itemize}


All experiments were conducted using the Computer Use framework with temperature set to 0. Our full system, Webscraper (Prompt + Tool), uses OpenAI GPT-o3-mini as the reasoning model within the Parse Tool. For both variants of our framework, our crawler-specific guidance is appended to the default system prompt. Each experimental run begins by launching a clean Firefox browser instance.

\subsection{Benchmark and Metrics}

Our benchmark comprises six mainstream Chinese and English news websites that exhibit diverse structures and interaction patterns, including infinite scroll, button-based pagination, and dynamic content loading. For each website, we constructed a ground truth dataset, or ``Golden'' set, by developing a manually-written, deterministic crawler. This Golden set contains the precise URLs, titles, and article content, serving as the reference for our evaluation.

While metrics like Exact Match are suitable for simple structured data, they are overly strict for long-form news articles, which often contain noisy elements like advertisements. To address this, we adopt ROUGE-L~\cite{lin-2004-rouge} as our primary metric for evaluating the quality of extracted titles and content. 
%
%
This metric offers three key advantages for our task: it penalizes both over-extraction (lowering precision) and under-extraction (lowering recall), while remaining robust to structural noise within the content.

To translate this score into a clear success indicator, we define a final binary metric called Correctness. An article is considered correctly extracted only if its URL is an exact match and both its title and content achieve a ROUGE-L score of at least 0.8.

We selected the 0.8 threshold based on two key considerations. First, while there is no universal academic standard, commercial systems like IBM's watsonx~\cite{IBM2024rouge} use 0.8 as a meaningful lower bound for ``high similarity.'' Second, our own empirical observations confirm this threshold: extractions with only minor noise consistently score above 0.8, whereas significant failures (e.g., partial extraction or LLM-generated text) score below 0.3. Thus, 0.8 serves as a reliable and well-grounded criterion for judging successful extraction.



\subsection{Preliminary Analysis: Experimental Stability}
To ensure the reliability of our results, we first conducted two preliminary analyses.
First, to determine an efficient number of experimental runs, we adopted a practical convergence criterion by analyzing the 95\% CI half-width for the performance metric across all nine experimental settings. 
Our analysis (Figure~\ref{fig:ci_convergence}) 
revealed a clear pattern of diminishing returns. 
The CI half-width decreased substantially up to the 30-run mark, after which the rate of precision gain slowed significantly. Balancing the negligible gain in precision against the significant computational cost, we concluded that 30 runs represent a robust and efficient standard for all experiments.
The CI half-width dropped sharply up to 30 runs, after which precision gains plateaued. Thus, 30 runs offer a robust and efficient standard for all experiments.

Second, we performed a temporal stability test by re-running experiments after a seven-day interval. The results (see Table ~\ref{tab:temporal_stability}) 
confirmed that the performance of our proposed method remained consistent, showing a variance of less than 5\% on all sampled websites.

\begin{figure}[htbp]
    \centering
    \includegraphics[width=0.8\columnwidth]{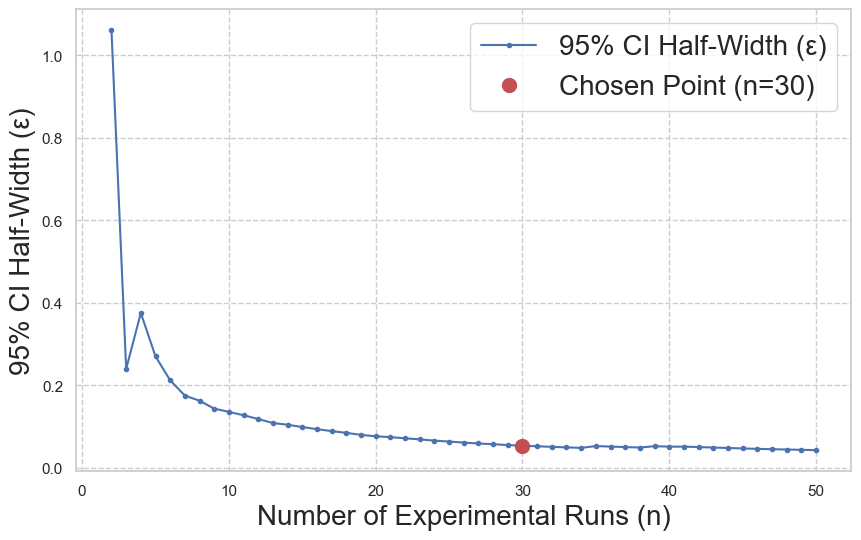}
    \caption{Convergence of the 95\% CI half-width for a representative scenario. ``Elbow point'' is visible around n=30, after which the curve flattens, indicating diminishing returns for additional runs.}
    \label{fig:ci_convergence}
\end{figure}

\begin{table}[htbp]
  \caption{Temporal stability results comparing accuracy at time T and T+7 days.}
  \label{tab:temporal_stability}
  \centering
  \begin{tabular}{l|cc|cc}
    \toprule
    & \multicolumn{2}{c}{Baseline} & \multicolumn{2}{c}{\makecell[c]{Webscraper \\ (Prompt + Tool)}} \\
    \cmidrule(lr){2-3} \cmidrule(lr){4-5}
    Website & T & T+7 & T & T+7 \\
    \midrule
  Website 3 & 0.103 & 0.061 & 0.511 & 0.533 \\
  Website 4   & 0.277 & 0.317 & 0.648 & 0.673 \\
  Website 5 & 0.145 & 0.179 & 0.820 & 0.820 \\
    \bottomrule
  \end{tabular}
\end{table}

\subsection{Main Results on News Websites}

\begin{figure}[t]
  \centering
  \includegraphics[width=\columnwidth]{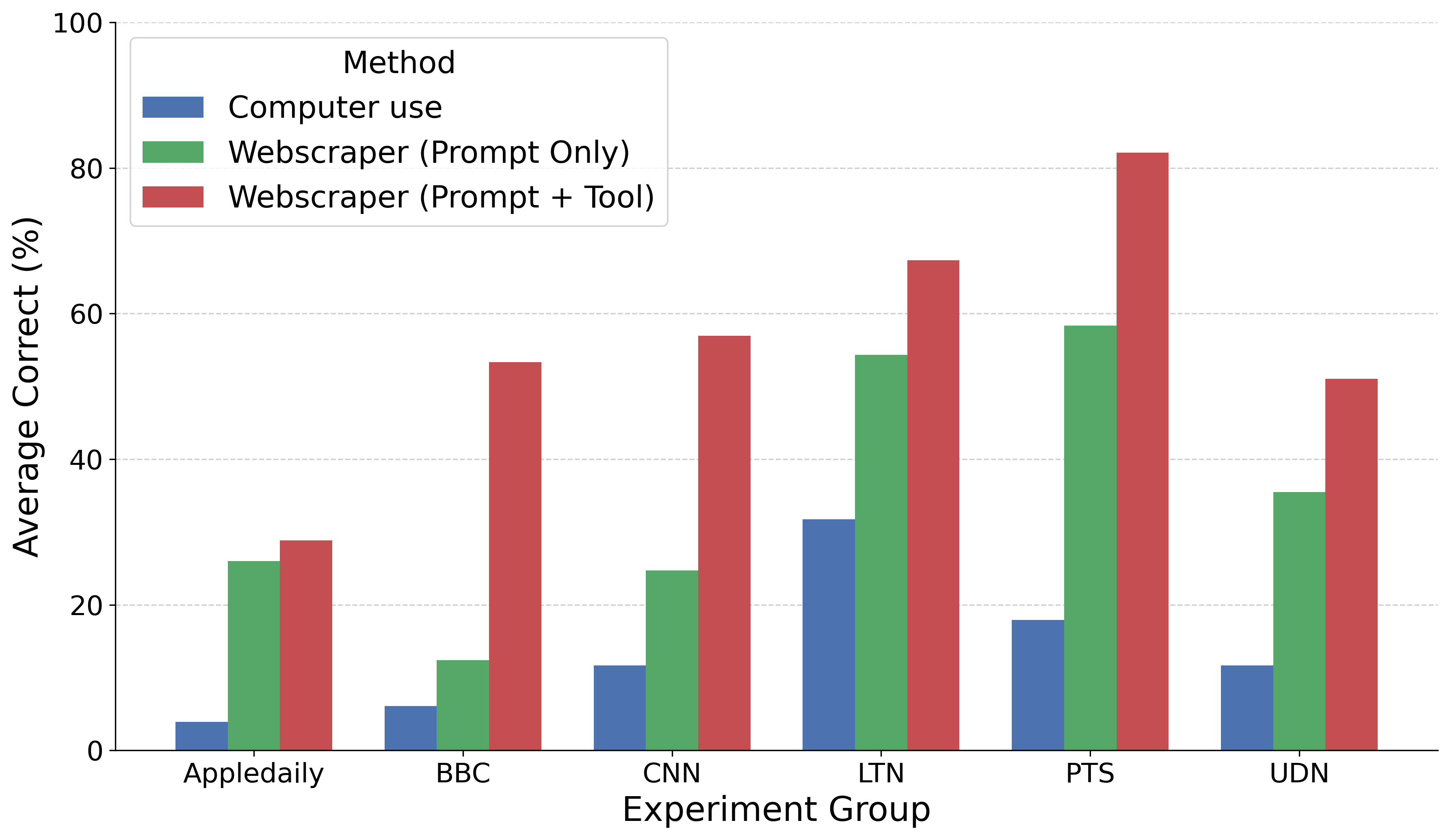}
  \caption{Performance comparison of our proposed methods against the baseline across six different news websites.}
  \label{fig:performance_comparison}
\end{figure}

Our primary experiments were conducted on the six news websites. The results are summarized in Figure \ref{fig:performance_comparison}.
As can be seen, our structured approach significantly outperforms the baseline.
Across all six websites, our full framework, Webscraper (Prompt + Tool), consistently and substantially outperformed the Baseline Agent. The baseline's success rate was particularly low (often below 50\%) on websites requiring multi-page navigation. The baseline agent struggled to identify and operate pagination mechanisms, succeeding only in rare instances (e.g., twice in 30 runs on LTN).
In contrast, our guided agent reliably handled these dynamic interactions. Even on tasks without pagination, our method proved superior by effectively using prompt-guided shortcuts (e.g., Ctrl+F) to locate specific content sections, a task the baseline struggled with.

Moreover, functional tools provide a distinct advantage over pure prompting.
When comparing Webscraper (Prompt + Tool) against the Webscraper (Prompt Only) ablation, the full framework achieved higher accuracy on every website. This result demonstrates that while a well-designed prompt provides crucial strategic guidance, equipping the agent with functional, specialized tools for parsing and merging leads to a tangible improvement in extraction success.

\subsection{Generalization to E-commerce Websites}

\begin{figure}[htbp]
    \centering
    \includegraphics[width=0.9\columnwidth]{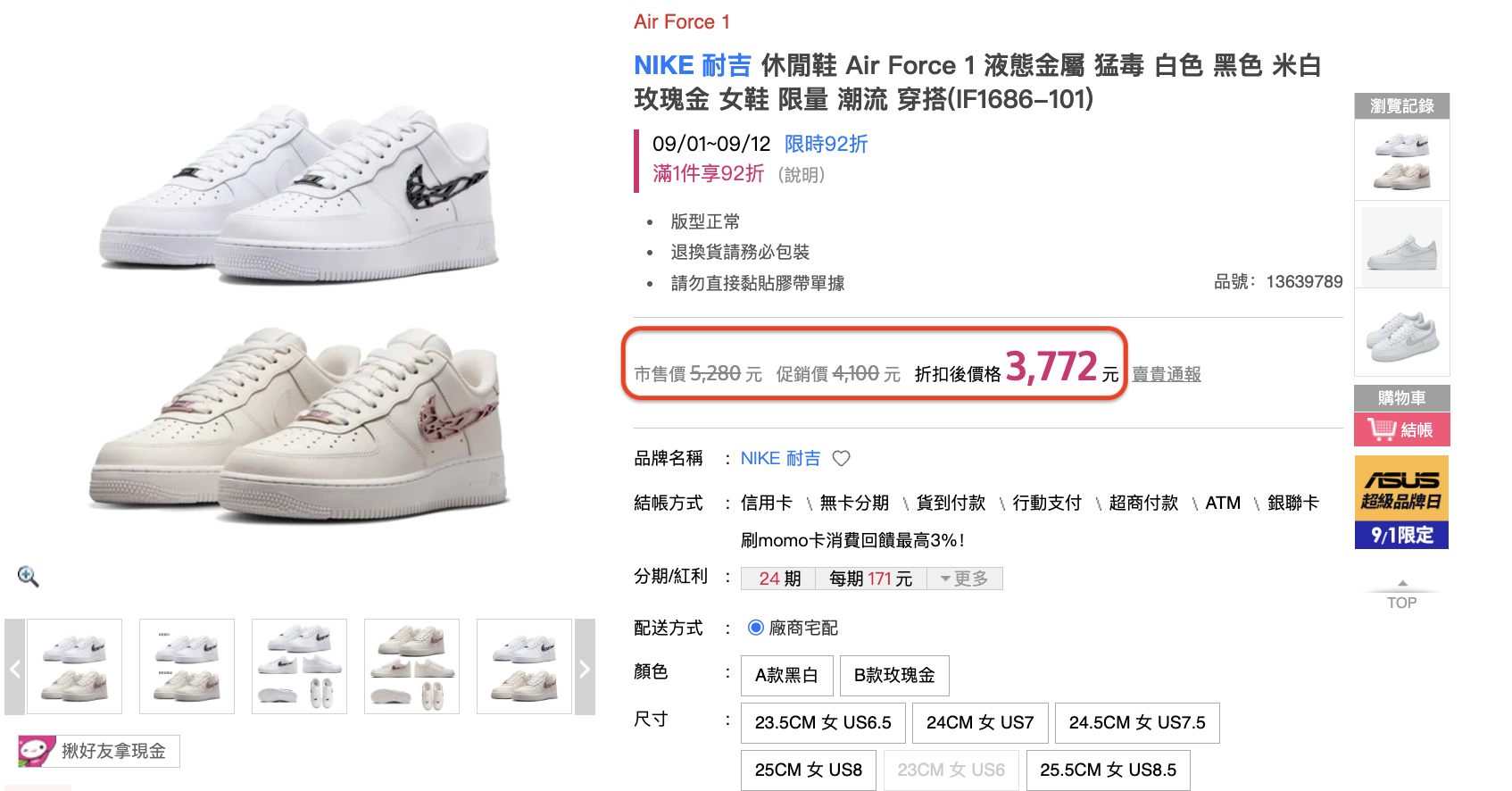}
    \caption{Extraction ambiguity on a Momo product page. The agent must identify the market price among multiple competing fields (e.g., promotional price, and discounted price, highlighted in the red box), which increases task complexity.}
    \label{fig:momo_challenge}
\end{figure}

For assessing the generalizability of our framework, 
we evaluated it on two large-scale e-commerce platforms, including Momo, Taiwan's largest e-commerce platform. The task involved extracting structured product data (e.g., prices, ratings), for which we used a stricter Correctness metric requiring a near-exact match.
The results are presented in Table ~\ref{tab:shopping_results}.  Again, our full framework (Webscraper (Prompt + Tool)) significantly outperformed both the prompt-only version and the baseline. The performance on Momo (0.242) was much lower than on Amazon (0.422).  This is due to the inherent ambiguity of the target page, which displayed multiple price fields simultaneously, making the extraction task more challenging. 
A screenshot of this challenging page is shown in Figure~\ref{fig:momo_challenge}.

The results  confirm the findings from our main experiments.
\begin{itemize}
\item Performance Ranking Holds: Our full framework (Webscraper (Prompt + Tool)) again significantly outperformed both the prompt-only version and the baseline, demonstrating that our approach is effective on different index-content style websites.

\item Task Complexity Matters: The performance on Momo (0.242) was lower than on Amazon (0.422). Our analysis revealed this was due to the inherent ambiguity of the target page, which displayed multiple price fields simultaneously, making the extraction task more challenging. A screenshot of this challenging page is provided in Figure~\ref{fig:momo_challenge}.
\end{itemize}

These results validate the general applicability of our method and underscore the consistent performance gains achieved by combining a guiding prompt with specialized tools.

\begin{table}[t]
  \caption{Performance comparison on two e-commerce websites. Best results are in bold.}
  \label{tab:shopping_results}
  \centering
  \begin{tabular}{lccc}
    \toprule
    Website & Baseline & \makecell[c]{Webscraper \\ (Prompt Only)} & \makecell[c]{Webscraper \\ (Prompt + Tool)} \\
    \midrule
    Momo & 0.000 & 0.040 & \textbf{0.242} \\
    Amazon & 0.027 & 0.138 & \textbf{0.422} \\
    \bottomrule
  \end{tabular}
\end{table}

\section{Discussion and Conclusion}

In this paper we proposed Webscraper, a framework that enhances a general-purpose, multimodal agent with specialized tools and a structured prompt to tackle dynamic web scraping. Our experiments on news and e-commerce sites demonstrated that this specialized approach significantly outperforms the baseline agent, particularly on complex websites requiring multi-page navigation.

A key finding is that general-purpose browsing agents, while proficient at multi-step tasks, are fundamentally inefficient for large-scale data extraction. SOTA agents like Browser Use~\cite{browser_use2024} adopt a sequential, one-page-at-a-time interaction strategy. However, this approach is unscalable, rapidly inflating the context window and leading to task failure on index pages with numerous links. In contrast, our method's ability to programmatically generate a single, reusable script for all content pages is far more efficient and robust for the widely used ``index-and-content'' web architecture. This highlights a crucial distinction: effective data extraction requires not just interaction capabilities, but a specialized, data-centric strategy.

Despite its strong performance, our approach is not without limitations, as revealed by our error analysis. The most critical failure modes occur in (1) complex navigation, where visual grounding can falter (e.g., inaccurate clicks after zooming), and (2) LLM-based code generation, where both our method and the baseline can produce buggy parsing scripts for inconsistent HTML structures. Furthermore, our framework is inherently limited to the index-and-content paradigm and would struggle with architectures like real-time WebSocket streaming or websites using advanced virtual scrolling, which unloads content from the DOM as the user scrolls.

We highlight two possible future directions to address these limitations.
First, to enhance robustness and reduce cost, the agent's successful interaction trajectory and validated code snippets could be compiled into a fully reusable, deterministic scraping script (e.g., using Selenium or Playwright). This would create a ``one-shot'' generation process, where the LLM is only needed once to create the scraper, eliminating the need for its intervention---and the associated costs and stochasticity---on subsequent runs.

Second, one may evolve the agent from a purely GUI-level operator to a more technically-aware analyst capable of observing both DOM mutations and network traffic. By monitoring real-time changes in the page's source code, the agent could learn to handle dynamic loading mechanisms like virtual scrolling. Simultaneously, by inspecting network requests and identifying protocols such as WebSockets, the agent could potentially bypass the GUI to tap directly into a website's underlying API endpoints. This hybrid approach, combining visual interaction with an understanding of the web's foundational architecture, may create a truly generalist and highly efficient web scraping agent.

\section*{Ethical Considerations}
Our research, which involves the automated scraping of publicly accessible websites, was designed with several ethical considerations in mind. We acknowledge that large-scale web scraping can raise concerns regarding copyright, server load, and the potential misuse of data. To mitigate these risks, we adopted the following principles:

First, our experiments were conducted exclusively on publicly available information. We did not attempt to bypass login walls, paywalls, or any other access control mechanisms. 

Second, to avoid imposing an undue burden on the websites' servers, all scraping tasks were executed with a low frequency and a significant delay introduced between requests. The total volume of data collected for our academic evaluation was minimal and represents a negligible fraction of the websites' normal traffic.

Finally, the data collected was used solely for the purpose of academic research—specifically, to evaluate the performance of our proposed framework. The data has not been and will not be redistributed or used for any commercial purposes. We believe that the insights gained from this research contribute to the understanding of autonomous agents, and we advocate for the responsible use of such technologies in the future.


\bibliographystyle{IEEEbib}
\bibliography{sample-base}

\appendix
\begin{remark}{
\section{ROUGE-L Formulation}
\label{sec:appendix_rouge}
ROUGE-L is based on the Longest Common Subsequence (LCS) between a candidate text $Y$ (extracted) and a reference text $X$ (golden). The recall ($R_{lcs}$), precision ($P_{lcs}$), and F1-score ($F_{lcs}$) are calculated as follows:
\begin{equation}
    R_{lcs} = \frac{\text{LCS}(X, Y)}{|X|}
    \label{eq:rouge_r_appendix}
\end{equation}
\begin{equation}
    P_{lcs} = \frac{\text{LCS}(X, Y)}{|Y|}
    \label{eq:rouge_p_appendix}
\end{equation}
\begin{equation}
    F_{lcs} = \frac{(1 + \beta^2) \cdot R_{lcs} \cdot P_{lcs}}{R_{lcs} + \beta^2 \cdot P_{lcs}}
    \label{eq:rouge_f_appendix}
\end{equation}
where $|X|$ and $|Y|$ are the lengths of the reference and candidate texts, respectively, and $\beta$ is set to 1.

}\end{remark}

\begin{remark}{
\section{Stability Analysis Details}
\label{sec:stability_analysis}


Figure~\ref{fig:appendix_cumulative_avg} shows that the performance metric tends to stabilize after approximately 30 runs

\begin{figure}[htbp]
  \centering
  \includegraphics[width=1\columnwidth]{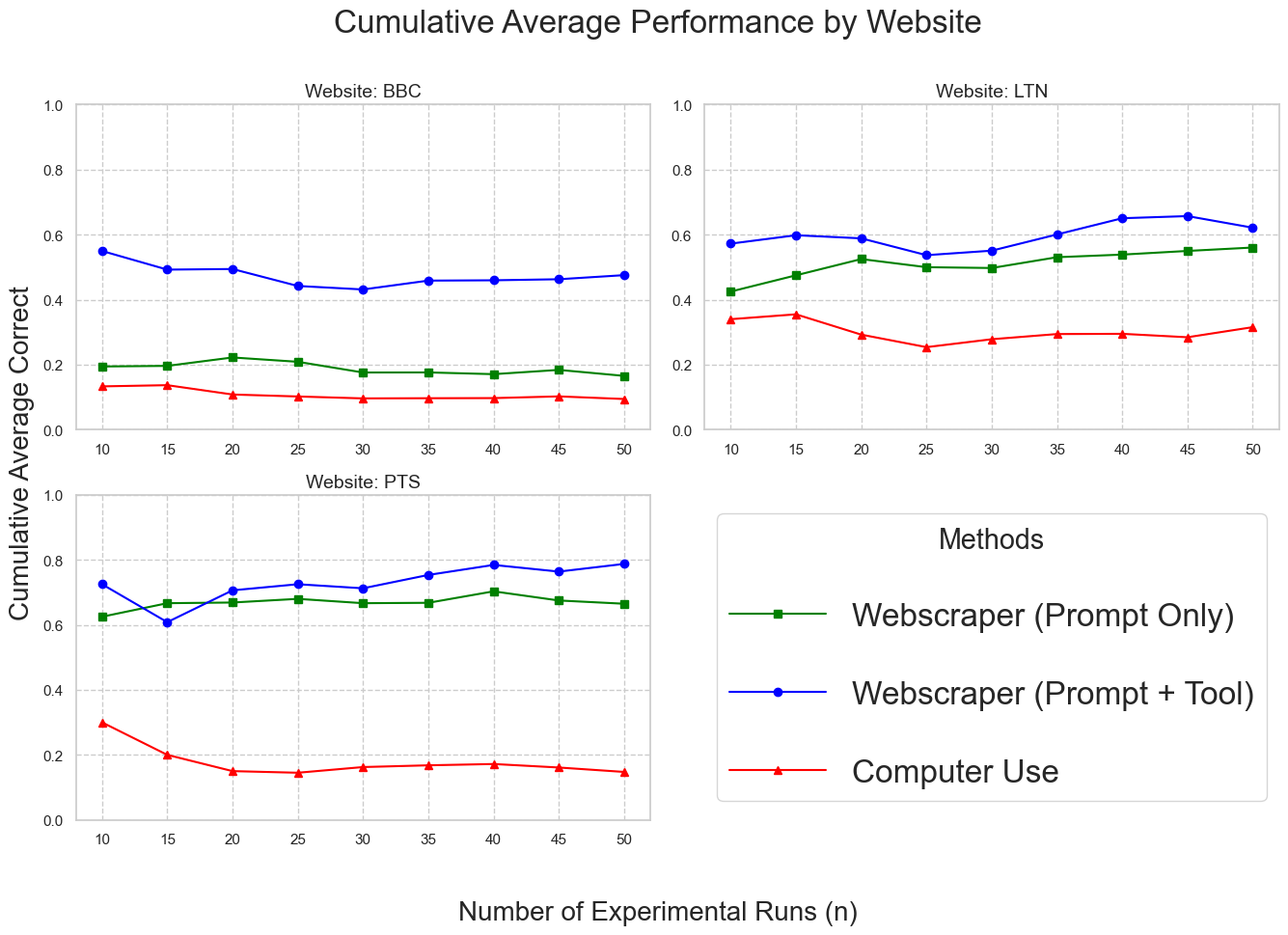}
    \caption{Cumulative average correctness plotted at 5-run intervals for a representative experimental setting. The plot shows that the performance metric tends to stabilize after approximately 30 runs, with subsequent fluctuations remaining minimal. This visual evidence supports our choice of n=30 as a stable and efficient number of runs.}
  \label{fig:appendix_cumulative_avg}
\end{figure}


Figure~\ref{fig:ci_convergence} illustrates a representative case where the metric stabilizes at run 30.

\begin{figure}[htbp]
    \centering
    \includegraphics[width=0.8\columnwidth]{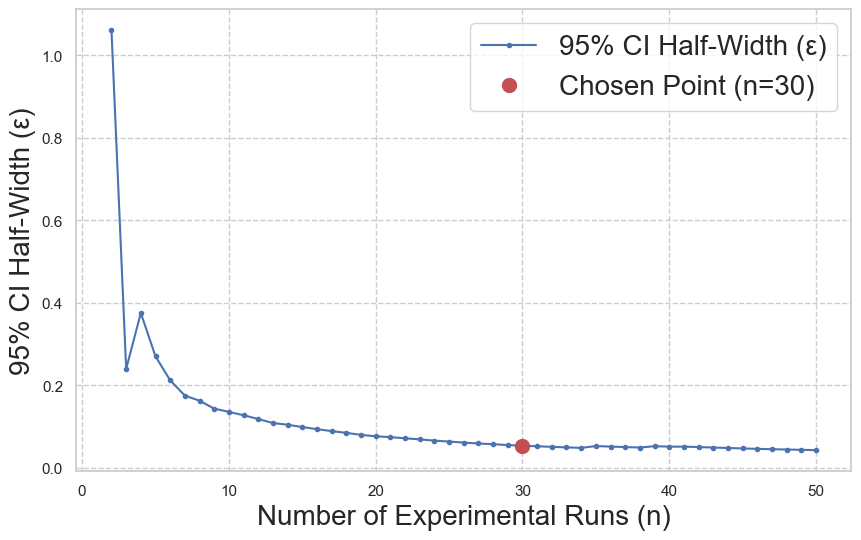}
    \caption{Convergence of the 95\% CI half-width for a representative scenario. ``Elbow point'' is visible around n=30, after which the curve flattens, indicating diminishing returns for additional runs.}
    \label{fig:ci_convergence}
\end{figure}


Table ~\ref{tab:temporal_stability} shows that our approach is stable across 7 days interval

\begin{table}[htbp]
  \caption{Temporal stability results comparing accuracy at time T and T+7 days.}
  \label{tab:temporal_stability}
  \centering
  \begin{tabular}{l|cc|cc}
    \toprule
    & \multicolumn{2}{c}{Baseline} & \multicolumn{2}{c}{\makecell[c]{Webscraper \\ (Prompt + Tool)}} \\
    \cmidrule(lr){2-3} \cmidrule(lr){4-5}
    Website & T & T+7 & T & T+7 \\
    \midrule
    BBC & 0.103 & 0.061 & 0.511 & 0.533 \\
    LTN & 0.277 & 0.317 & 0.648 & 0.673 \\
    PTS & 0.145 & 0.179 & 0.820 & 0.820 \\
    \bottomrule
  \end{tabular}
\end{table}

 }\end{remark}

\begin{remark}{
\section{Extraction Challenges on Momo E-commerce Platform}
\label{sec:appendix_momo}

Our analysis indicates that the lower scraping performance on the Momo e-commerce platform is primarily due to the ambiguity of its product pages.
Figure~\ref{fig:momo_challenge} illustrates a representative case that highlights this challenge.

}\end{remark}

\end{document}